\documentclass[conference]{IEEEtran}
\IEEEoverridecommandlockouts

\usepackage{cite}
\usepackage{amsmath,amssymb,amsfonts}
\usepackage{algorithmic}
\usepackage{graphicx}
\usepackage{textcomp}
\usepackage{xcolor}
\def\BibTeX{{\rm B\kern-.05em{\sc i\kern-.025em b}\kern-.08em
    T\kern-.1667em\lower.7ex\hbox{E}\kern-.125emX}}

\usepackage{cite} 
\usepackage{xcolor} 
\usepackage{amssymb,graphicx,color} 
\usepackage{latexsym} 
\usepackage{gensymb} 
\usepackage{subcaption} 
\usepackage{hyperref} 

\begin{document}

\title{Data-Driven Socio-Economic Deprivation Prediction via Dimensionality Reduction: The Power of Diffusion Maps\\
}

\author{\IEEEauthorblockN{June Moh Goo}
\IEEEauthorblockA{\textit{Department of Civil, Environmental and Geomatics Engineering} \\
\textit{University College London}\\
London, UK \\
june.goo.21@ucl.ac.uk}
}

\maketitle

\begin{abstract}
This research proposes a model to predict the location of the most deprived areas in a city using data from the census. Census data is very high-dimensional and needs to be simplified. We use the diffusion map algorithm to reduce dimensionality and find patterns. Features are defined by eigenvectors of the Laplacian matrix that defines the diffusion map. The eigenvectors corresponding to the smallest eigenvalues indicate specific characteristics of the population. Previous work has found qualitatively that the second most important dimension for describing the census data in Bristol, UK is linked to deprivation. In this research, we analyse how good this dimension is as a model for predicting deprivation by comparing it with the recognised measures. The Pearson correlation coefficient was found to be greater than 0.7. The top 10 per cent of deprived areas in the UK, which are also located in Bristol, are extracted to test the accuracy of the model. There are 52 of the most deprived areas, and 38 areas are correctly identified by comparing them to the model. The influence of scores of IMD domains that do not correlate with the models and Eigenvector 2 entries of non-deprived Output Areas cause the model to fail the prediction of 14 deprived areas. The model demonstrates strong performance in predicting future deprivation in the project areas, which is expected to assist in government resource allocation and funding greatly. The codes can be accessed here:    \href{https://github.com/junegoo94/diffusion_maps}{https://github.com/junegoo94/diffusion\_maps}.
\end{abstract}


\section{Introduction}
\label{introduction}
The definition of Socio-Economic deprivation is the lack of necessities to participate in such as poverty, lack of access to jobs, and exclusion from services \cite{gordon2000poverty}. Socio-economic deprivation is, of course, combined in specific geographic areas. Areas with a relatively high demand and a lower supply of necessities, compared to other regions, should be socio-economically deprived areas \cite{sarkar2014patterns}. The government has considered solutions to improve the quality of life in deprived areas. Therefore, it is important to measure deprivation depending on how much of those areas are suffering from deprivation to allocate resources efficiently. In the past, necessities were limited to the indicators of food, shelter and warmth. Therefore, reducing deprivation was focused on these three factors. However, considering only those factors cannot reflect appropriate socio-economic development because those factors are too vague and not detailed enough to measure various types of deprivation.

In order to overcome the shortcomings of past methods and identify the patterns of socio-economic deprivation, many more specific indicators are now used. Factors such as income (economic factors), education and health have been added to measure the deprivation \cite{sarkar2014patterns}. Depending on the study \cite{2015deprivation}, the selected indicators are changed by adding or removing indicators such as age, gender and employment \cite{sarkar2014patterns}. 
The Department for Communities and Local Government (DCLG) \cite{2015deprivation} has published a statistically analysed report on deprivation every 3 to 5 years. This report measures the current level of deprivation in each region. However, it is difficult to obtain information about how the level of poverty will change in the future. 
Previous work \cite{barter2018manifold, xiu2024mobility, ghafourian2020mathematical} has suggested that the census data, which had a common point of looking at changes in the amount of population measurement, will be related to the prediction of deprivation. Therefore, we use the census data as original data and transform it through the technical process to select the characteristics that correlate with the deprivation data.

\begin{figure}[h]
    \centering
    \includegraphics[width=1\linewidth]{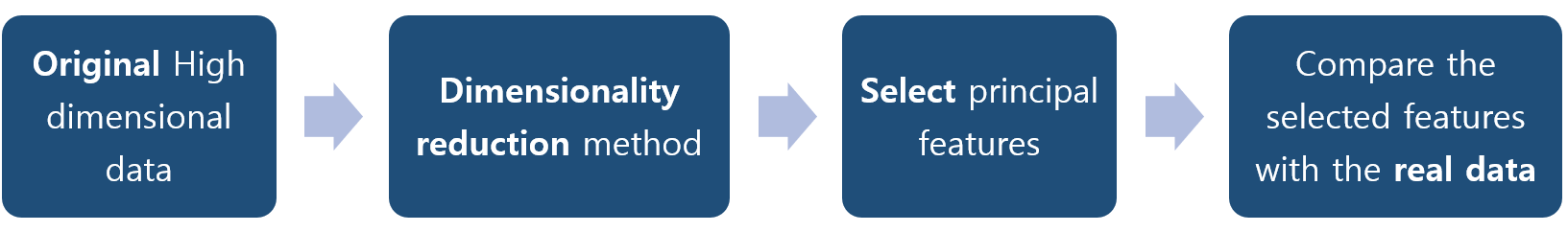}
    \caption{The main steps for technical process}
    \label{fig:Intro}
\end{figure}

The technical process of this project will be primarily divided into four steps, as shown in Fig.\ref{fig:Intro}. Since the census data (original data) is a high-dimensional dataset, we want to obtain the principal features of such a data set. Hence, we use a dimensionality reduction technique. In this project, we will use a diffusion map, one of the dimensionality reduction methods \cite{coifman2006diffusion}. The diffusion map technique is a way to understand a high-dimensional dataset embedded in the non-linear manifold and select the features relevant to our interest. In other words, the diffusion maps can relate the spectral properties from the diffusion map algorithm to the geometry of the data. From the models constructed by the diffusion map algorithm, we will evaluate the accuracy and possibility of developing this model for deprivation. The results of this research have the potential to support the government or local government policy on deprivation, for example, to allocate or adjust resources for deprived areas.

\section{Backgrounds and related works}
\label{literature_review}
\subsection{UK Census data}
The UK Census is collected every 10 years (we use data from 2001 and 2011 in this project) in order to survey the population. The data itself is published 3 years later by the Office for National Statistics(ONS), and are used, for example, to plan the appropriate allocation of resources\cite{modern}. Census data are published not at an individual level but aggregated into geographical areas. In this paper, we particularly focus on Output Areas (OA), the geographical levels, and Super Output Areas (SOA)\cite{censusgeography}.

The OAs were built from postcode units for the statistical purposes of the census. OAs for the 2001 census in the UK (except Scotland) are designed to have a similar population size and to be as socially homogeneous as possible, in terms of ownership of households and type of dwelling \cite{censusgeography}. For example, an OA cannot include both urban areas and rural areas, since each OA should consist of either urban or rural postcodes. The shape of each 2001 OA is similar, and their boundaries are usually constrained by the major roads. The threshold of 2001 OA size has been fixed as a minimum of 40 resident household and 100 resident people\cite{censusgeography}.

The OAs are used for Census output. They are also the basis of Super Output Areas (SOAs) which are stable and consistent in the size of the areas. SOAs were used to calculate the deprivation in 2004 and have become standard areas for the Indices of Deprivation\cite{local}.
SOAs have two tiers: Lower Layer Super Output Areas (LSOAs) and Middle Layer Super Output Areas (MSOAs). LSOAs typically contain 4 to 6 OAs, with a population of around 1500, while MSOAs have a population of around 7200. For the analysis of the 2011 Census, the ONS tried to maintain the stability of SOA and its two tiers as far as possible with the 2001 Census \cite{censusgeography}. However, if the population or the number of households changed significantly, the LSOAs and MSOAs were modified from the 2001 Census. Simplistically, certain LSOAs or MSOAs where populations were too big or small were either split into 2 or more areas or merged \cite{censusgeography}. For example, the total number of OAs and LSOAs in 2001 was 181,287 and 34,378 respectively, and in 2011 have changed to 181,408 and 34,753 for England and Wales \cite{censusgeography}.

In this project, we considered the local authorities of Bristol, North Somerset, Bath, North East Somerset, and South Gloucestershire, where we have local knowledge that allows us to understand and interpret the results. These regions provide a diversity of the census by combining the data of urban and rural areas and including all of the Bristol urban areas. This combined area of Bristol and its surrounding Local authorities contains 3490 OAs and 667 LSOAs.

\subsection{English Indices of Deprivation}
\label{IMD_description}
Classifying areas by deprivation level is important as the government or local government can choose which areas to allocate resources or provide funding. Hence, the amount of resources allocated or funded depends on how to measure the deprivation. Thus, since the 1970s, the Department for Communities and Local Government (DCLG) has collected local deprivation data and set the indexes of deprivation based on the census in England \cite{2010deprivation,2015deprivation}. These census-based deprivation indices are important uses of social statistics, as they construct key criteria in the allocation of resources to local governments \cite{kawachi2003neighborhoods}. Unfortunately, in the UK Census data, no questions were asked about income, nor to investigate either poverty or deprivation specifically. Therefore, DCLG released English deprivation indices separately from the census, which focuses only on poverty and deprivation. The data are based on LSOAs, one of two tiers of SOA, which is a suitable standard area for investigating the indices of deprivation \cite{2010deprivation,2015deprivation}.

DCLG used 38 indicators (37 indicators in 2015) to measure the deprivation of each LSOA, organised into 7 distinct domains \cite{2010deprivation}. The deprivation of each LSOA is measured by scores and ranks of the 7 domains, appropriately weighted. The 7 domains are as follows:
\begin{itemize}
    \item Income Deprivation (22.5\%)
    \item Employment Deprivation (22.5\%)
    \item Health Deprivation and Disability (13.5\%)
    \item Education skills and Training Deprivation (13.5\%)
    \item Barriers to Housing and Services (9.3\%)
    \item Living Environment Deprivation (9.3\%)
    \item Crime (9.3\%)
\end{itemize}
These 7 domains are combined into a single Index of Multiple Deprivation (IMD) by the weights of each domain. The Index of Multiple Deprivation (IMD) is widely used for various purposes, for example by national and local governments to identify places for prioritising resources and funding. It also supports local growth and gives more evidence for setting a range of local strategies and service planning. Furthermore, IMD is used in bids for funding based on nationally comparable measures of deprivation \cite{2015deprivation}.

However, there are cautions when observing this index. First, since the indices show the relative deprivation along the LSOAs in one particular year, it is not a good way to compare the deprivation score of an LSOA in different years. Second, all people who live in the most deprived areas are not necessarily deprived; similarly, deprived people can live in non-deprived areas \cite{2015deprivation}.

Therefore, we build models that represent the deprivation level of LSOAs. These models can be a preview of how the deprivation level of LSOAs change in the future. Furthermore, we will look at the deprivation on the OA level to notice the deprivation more specifically on the smaller spatial scale.

\subsection{Dimensionality Reduction}
Real-world data such as that we use in our research can often be expressed in high dimensional space, with one dimension per category. This makes finding patterns in the data a significant conceptual and computational challenge. With census data existing in a 1450-dimensional space, it is necessary to reduce dimensionality using techniques like manifold learning \cite{barter2018manifold}. The main idea of dimensionality reduction is transforming a high dimensional dataset into a low dimensional space, which can represent significant properties of the data \cite{van2009dimensionalityreduction,barter2018manifold}. Widely-known dimensionality reduction techniques include Principal Components Analysis (PCA), factor analysis, and classical scaling \cite{van2009dimensionalityreduction}.

{\bfseries Principal Components Analysis (PCA)} is a popular technique for dimensionality reduction. It works by embedding the data into a linear subspace of lower dimension \cite{jolliffe2011principal}. PCA builds a low-dimensional manifold of the data that describes the largest possible variance, so the principal component contains as much of the variability in the data as possible. PCA has been applied to a large number of problems in biology \cite{bonnier2012understanding}, finance \cite{janicijevic2022principal}, economics \cite{chen2010measuring} and plenty of search engines \cite{forbes2005efficient}. However, it has two main weaknesses. First, PCA can find only linear structures, such as straight lines or flat planes in the data  \cite{barter2018manifold}. Thus, if the data is linearly correlated, then PCA is a good method to find the directions representing the dataset. Otherwise, it cannot capture the direction in a non-linear data set. Second, PCA focuses mainly on maintaining large pairwise distances, which means it considers all distances of every data point pair, even though maintaining small pairwise distances is much more important in finding the similarity of the data point pairs. Similarly, other linear techniques, such as factor analysis and classical scaling, have drawbacks similar to PCA, so they are not an ideal method to solve the limitations above \cite{barter2018manifold,van2009dimensionalityreduction}.

However, high-dimensional data often has a non-linear manifold. Non-linear manifolds are the standard mathematical objects which is not possible to represent the data on the linear hyperplane. An example of a non-linear manifold is the torus. Thus, non-linear techniques offer advantages over linear techniques since most of the real-world data tend to form a non-linear manifold. Coifman et al. \cite{coifman2006diffusion} introduced the \noindent{\bfseries Diffusion Map}, which finds meaningful geometric descriptions of data sets regardless of distant pairs, to overcome the shortcomings of linear techniques described above. The first step for the diffusion map algorithm is to embed data points into Euclidean space and calculate the distances (Euclidean distances) between all pairs of nodes since distances describe the relationship between data pairs. Then, the data pairs which have less relationship are discarded. In this case, the distances of far data pairs are excluded in terms of their connectivity. To drop the distances too far to be considered, a heuristic procedure is used in our paper (See Section \ref{methodology}) \cite{coifman2006diffusion}. Visualising the network of this dataset\cite{barter2018manifold} is possible. Each data point is represented as a node, and the weighted links indicate the proximity of data points. This network can be expressed as a Laplacian matrix, which is a matrix representation of a graph. The Laplacian matrix can be used to find useful properties of the network \cite{godsil2013algebraicLaplacian}; for example, its eigenvectors and eigenvalues can locate manifolds and the structure of the network because eigenvalues of the Laplacian matrix represent the connectedness of the network \cite{marsden2013eigenvalues}. The eigenvectors with the smallest positive eigenvalues are selected because they are the main direction of manifolds with large variation \cite{barter2018manifold}. 

\section{Methodology}
\label{methodology}
\begin{figure*}[!t]
    \centering
    \includegraphics[width=0.85\textwidth]{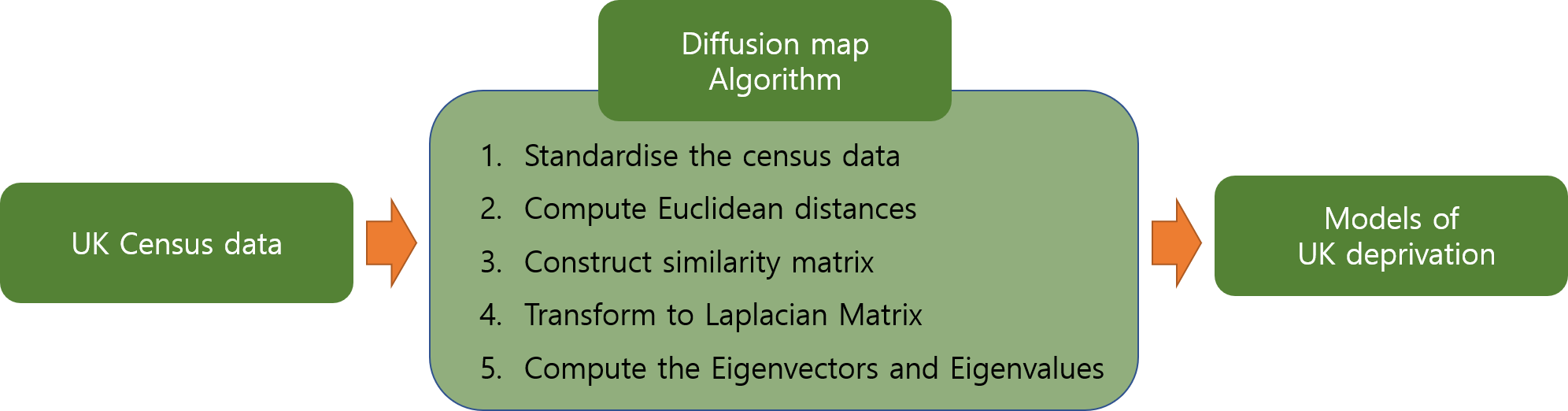}
    \caption{Overall technical processes of the our approach}
    \label{fig:algorithm}
\end{figure*}
This section provides an in-depth overview of our approach using a diffusion map. In this project, we applied the diffusion map algorithm through a structured five-step process, as illustrated in Fig. \ref{fig:algorithm}. The procedure begins with the standardization of the dataset, followed by constructing a similarity matrix, which is then transformed into a Laplacian matrix. We first construct two diffusion maps: the OA diffusion map and the LSOA diffusion map. OA diffusion map is the model that uses census data as an input (The census data is OA-based collection). However, as the UK deprivation data is based on LSOA boundaries, it is an essential step to transform the UK census data, which is collected with OA geographical resolution, into LSOA geographical resolution. We can now construct an LSOA diffusion map model, which uses the census data expressed by LSOAs. Therefore, before we start applying the census data to build the LSOA diffusion map, we re-organise the dataset to make the census data at the level of LSOA. We grouped all OAs in a given LSOA to transform the data by averaging the census data entries. As a result, two datasets are constructed, OA-based and LSOA-based census data, each with its diffusion map.

We begin with the census data organised in two matrices, 
$A_{OA}$ and $A_{LSOA}$, where each matrix has dimensions 
\textit{M×N}. In this setup, $M=3490$ for OAs and $M=667$ for LSOAs, while $N=1450$ represents the number of census variables. To enable a consistent comparison across all variables, we standardise each column in matrix $A$, and each census variable is normalised to have a mean of zero and a standard deviation of 1. This normalization ensures that all variables are on the same scale, allowing us to consider them equally in the analysis. As a result, the components of the normalised matrix \textbf{B} are

\begin{equation}
       B_{ij} = \frac{A_{ij}-\mu_{j}}{\sigma_{j}},
\end{equation}
where\\
\begin{equation}
       \mu_{j} = \dfrac{\sum\limits_{i=1}^{\ M} A_{ij}}{\textit{M}},
\end{equation}
and\\
\begin{equation}
    \sigma_{j}=\sqrt{\dfrac{\sum\limits_{i=1}^{\ M} (A_{ij}-\mu_{j})^2}{\textit{M}}}.
\end{equation}
Then, we compute Euclidean distances between data points of the standardised matrix \textbf{B}. So we define the $M\times M$ Euclidean distances matrix as a matrix \textbf{D} with entries,\\
\begin{equation}
    D_{ij}=\sqrt{\sum\limits_{k=1}^{\ N} (B_{ik}-B_{jk})^2},
\end{equation}
where \textit{N} = 1450. $D_{ij}$ indicates the Euclidean distance between geographical output areas $i$ and $j$.\\\\
We then convert the entries of the matrix \textbf{D} in order to indicate the similarity scores of each pair of data points. The idea of this step is to transform the distances, $D_{ij}$, to $1/D_{ij}$ for $i \neq j$, to construct a $M\times M$ similarity matrix \textbf{C}, with entries.\\
 \begin{equation}
       C_{ij} = 
        \begin{cases}
            \dfrac{1}{D_{ij}}, & \text{if $i \neq j$,} \\
\\
            0 & \text{otherwise.}
        \end{cases}
    \end{equation}
Therefore, a large value of $C_{ij}$ indicates close similarities between the geographical standards (OA or LSOA).

Then, we threshold the entries of the matrix \textbf{C} and set 'small' entries to zero. It is hard to define the benchmark whether the entries are 'small'. Thus, we keep the value of $C_{ij}$, if $C_{ij}$ is in the top 10 highest similarity scores of either $OA_{i}$ ( or $LSOA_{i}$) or $OA_{j}$ (or $LSOA_{j}$). Otherwise, it is set to zero. Hence, each node has at least 10 links. This step removes the low-similarity scores and gives the benefit of leaving essential links, numerically more efficient \cite{barter2018manifold}.

With the threshold matrix \textbf{C}, we can construct the Laplacian matrix, which can then be used to find useful local properties of the network. We define the normalised $M\times M$ Laplacian matrix, \textbf{L} with entries \cite{barter2018manifold}
    \begin{equation}
       L_{ij} = 
        \begin{cases}
            1, & \text{if $i = j$,} \\
\\
            -\dfrac{C_{ij}}{\sum\limits_{k=1}^{M} C_{kj}} & \text{otherwise.}
        \end{cases}
    \end{equation}
We compute eigenvalues $\lambda$ and eigenvectors $v$ of \textbf{L} to investigate the structure of the dataset. They are defined by the equation
\begin{equation}
       \textbf{L}\textit{v} = \lambda\textit{v}.
\end{equation}
Laplacian matrices are symmetric and semi-definite, and so their eigenvalues are larger or equal to zero. The number of components of data points in the network is the number of zero eigenvalues \cite{barter2018manifold}. Therefore, there exists at least one eigenvalue which is zero. However, eigenvectors with zero eigenvalues do not carry out any matrix features, so we do not consider them. Hence, the eigenvalues we are interested in are the smallest non-zero positive eigenvalues as they represent the main directions with a large variance of the manifolds in the data. We select the two eigenvectors with the first two smallest eigenvalues of each of the two datasets: the OA and LSOA datasets. As these eigenvectors assign a value to each OA and LSOA, we can now visualise the eigenvectors on the spatial map \cite{barter2018manifold}.

\section{Visualization of Diffusion Maps}
\label{visualization}
\subsection{LSOA diffusion map}

\begin{figure}[ht]
    \centering
    \begin{subfigure}[b]{0.4\linewidth}
        \centering
        \includegraphics[width=\linewidth]{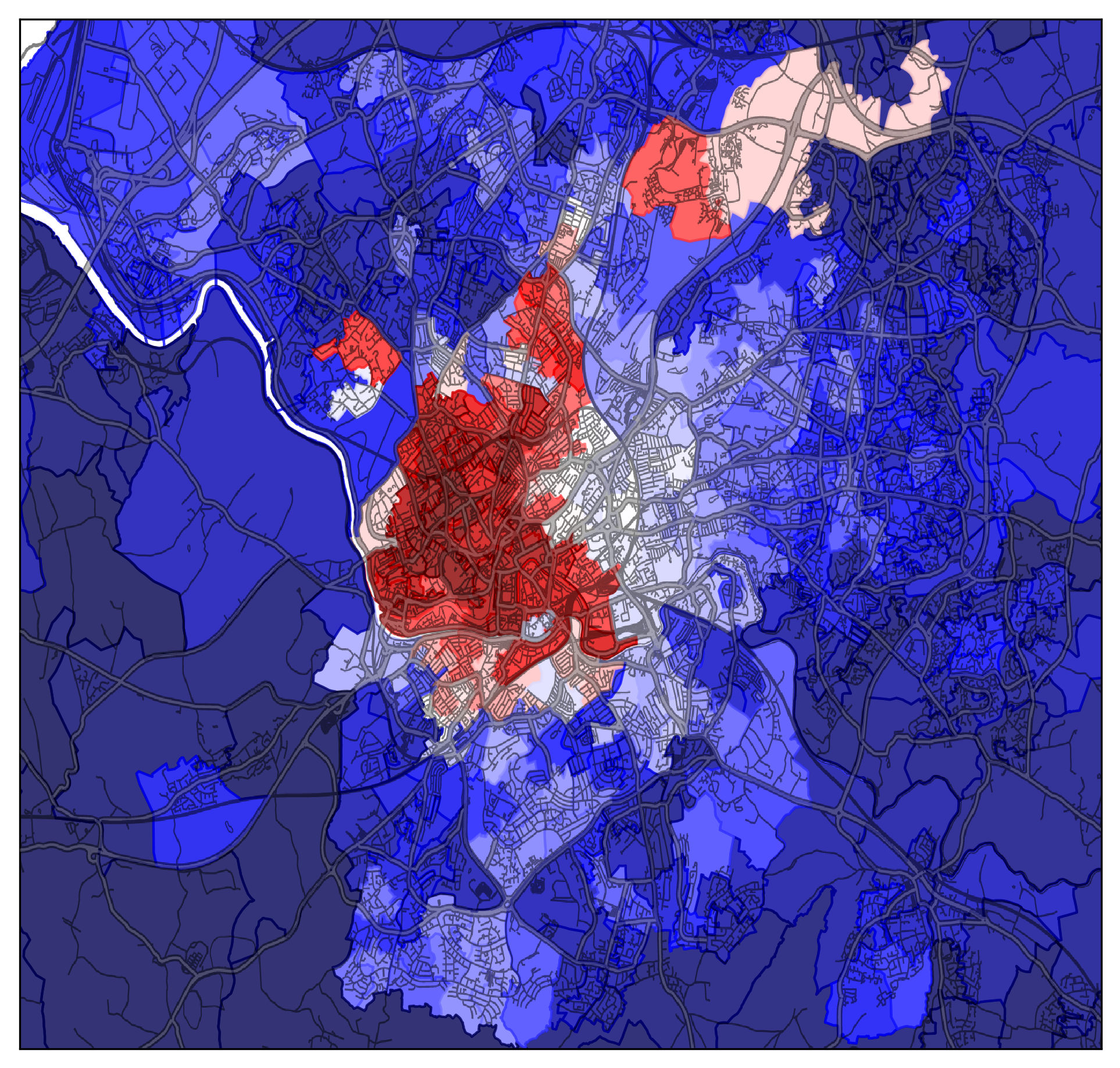}
        \caption{Eigenvector 1}
        \label{fig:lsoa1}
    \end{subfigure}
    \begin{subfigure}[b]{0.4\linewidth}
        \centering
        \includegraphics[width=\linewidth]{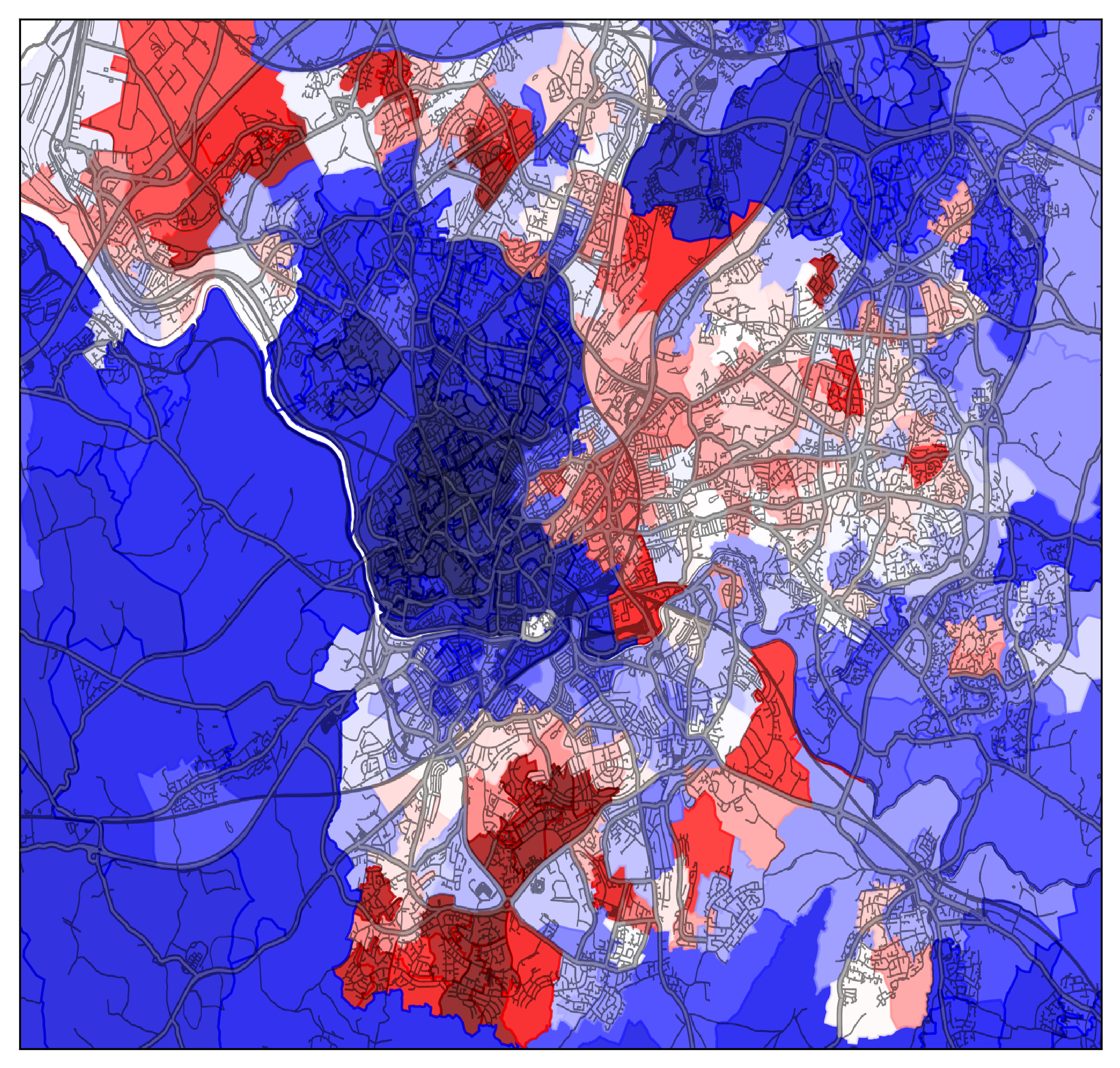}
        \caption{Eigenvector 2}
        \label{fig:lsoa2}
    \end{subfigure}
    \begin{subfigure}[b]{0.084\linewidth}
        \centering
        \includegraphics[width=\linewidth]{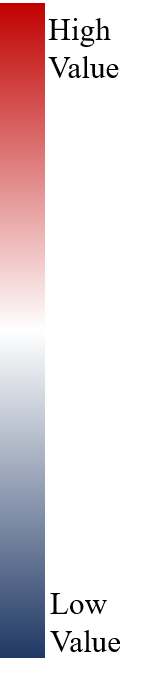}
        \label{fig:scalebar1}
    \end{subfigure}
    \caption{Selected eigenvectors of LSOA diffusion map}
    \label{fig:lsoadiff}
\end{figure}

The LSOA diffusion map results from the diffusion map algorithm when applied to the census data at the LSOA level. As mentioned above, the eigenvectors of the Laplacian matrix corresponding to the smallest positive eigenvalues span the main direction of the manifolds in the data. Each component of the eigenvectors corresponds to an LSOA, which means each eigenvector has spatial information. To visualise the spatial information of each eigenvector, we can plot a map of the LSOAs, coloured by the eigenvector component. Fig. \ref{fig:lsoadiff} shows coloured maps of the eigenvectors with the two smallest positive eigenvalues where red-colored areas have high-value entries, and blue-colored areas have low-value entries of the eigenvector. Each eigenvector represents its pattern of the areas in the map. From Eigenvector 1, shown in Fig. \ref{fig:lsoa1}, for example, the red areas indicate where the student population is high, and blue where the student population is low \cite{barter2018manifold}.\\\\
The Eigenvector 2, corresponding to the second smallest eigenvalue, is shown in Fig. \ref{fig:lsoa2}. We see complex patterns of areas. We notice this pattern of areas correlates with the location of major council estates. In the UK, local governments have provided houses, known as council housing. These houses are given to the homeless, people who live in cramped conditions and need to move because of a disability, medical, welfare or hardship reasons. While the income and savings of the individuals are above a certain amount, the individuals are excluded from the scope of the offer. When all conditions are put together, Eigenvector 2 can be considered an indicator of socio-economic deprivation. 

\subsection{OA diffusion map on LSOA}
The diffusion map in Fig. \ref{fig:lsoadiff} is built based on LSOA boundaries. We construct another diffusion map based on the original census data at the OA level. 3490 OAs exist in Bristol and surrounding areas; thus, the number of entries in each column in the OA diffusion map is 3490. As mentioned in Section \ref{IMD_description}, UK deprivation data are collected with LSOA boundaries. Therefore, a method for fitting the OA diffusion map into LSOA boundaries is necessary to compare which diffusion map is a better model for UK deprivation data. The idea is to allocate a new entry for certain LSOA by computing the average of the OA Eigenvector entries located in the boundary (Refer to \eqref{eq:LSOA2OA}).

\begin{equation}
    N_{ij}=\frac{1}{n}\times\sum^{n}_{k=1}(E_{kj})
    \label{eq:LSOA2OA}
\end{equation}
where:
\begin{itemize}
    \item \textit{E}: is an eigenvector of the OA diffusion map,
    \item \textit{N}: is the new 'eigenvector' at LSOA level,
    \item \textit{n}: is the number of OAs located in the certain $LSOA_{i}$
\end{itemize}

We now obtain 667 new entries in each 'eigenvector' column, which makes plotting spatial information at the LSOA level possible. The entry values are based on the OA diffusion map, so differences exist between the LSOA diffusion map and the OA diffusion map with LSOA boundaries. Therefore, we investigate these differences by exploring the patterns of areas with the new diffusion map.

\begin{figure}[h]
    \centering
    \begin{subfigure}[b]{0.4\linewidth}        
        \centering
        \includegraphics[width=\linewidth]{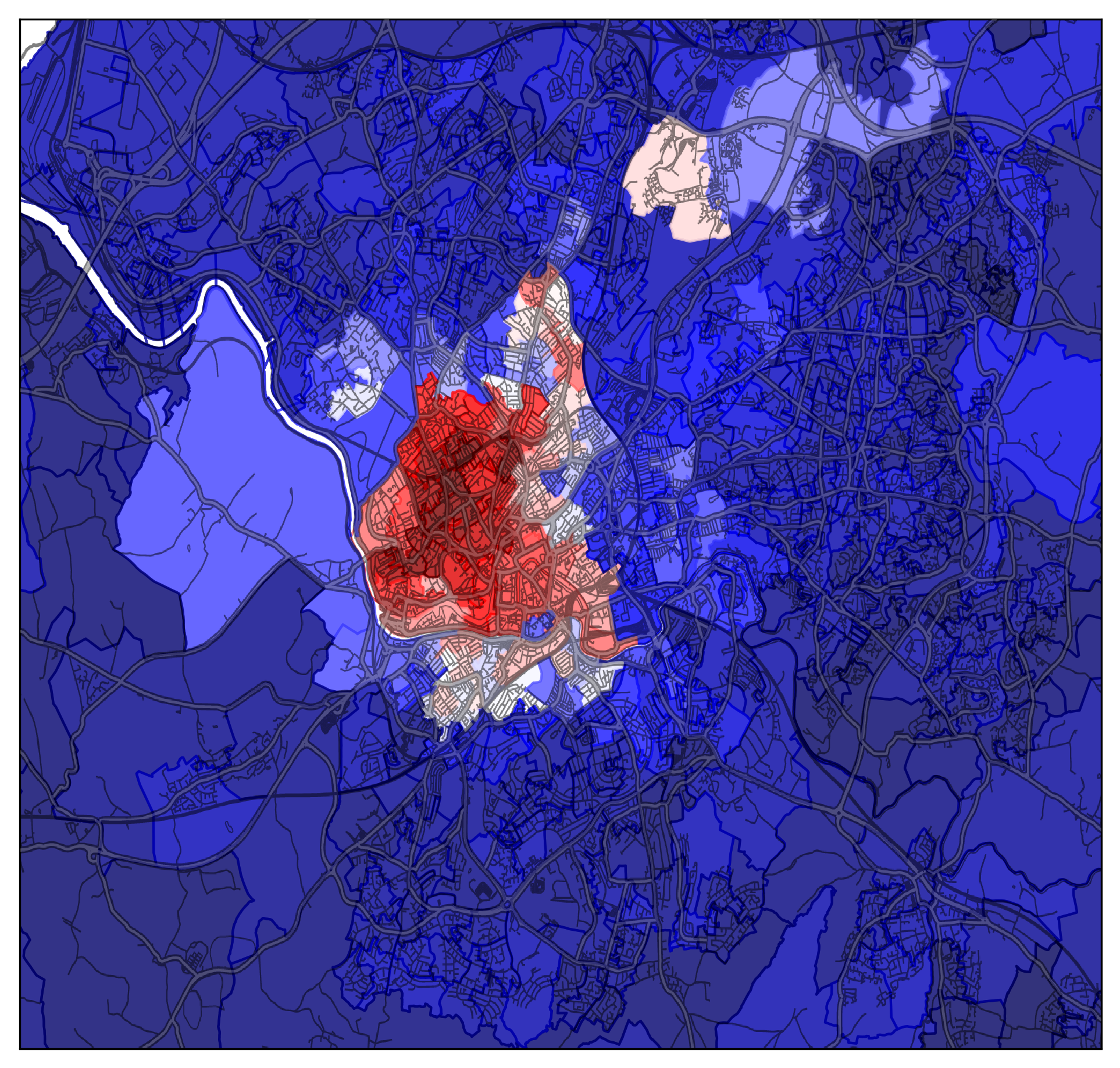}
        \caption{Eigenvector 1}
        \label{fig:oa1}
    \end{subfigure}
    \begin{subfigure}[b]{0.4\linewidth}        
        \centering
        \includegraphics[width=\linewidth]{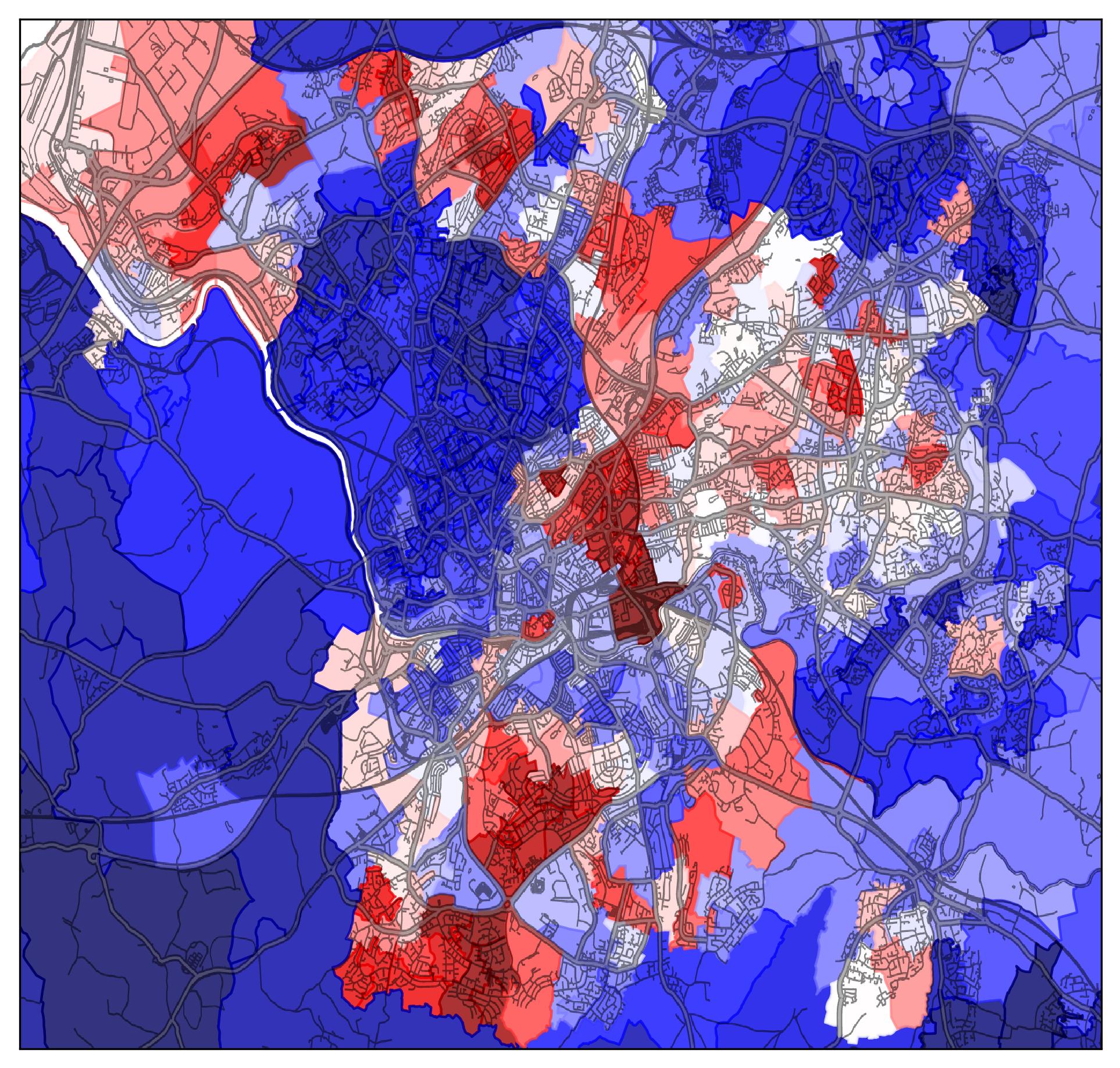}
        \caption{Eigenvector 2}
        \label{fig:oa2}
    \end{subfigure}
    \begin{subfigure}[b]{0.084\linewidth}        
        \centering
        \includegraphics[width=\linewidth]{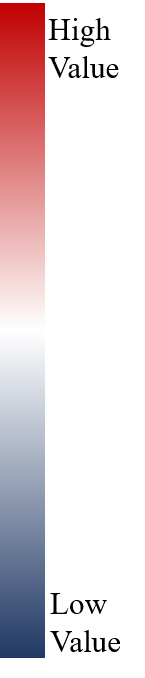}
        \label{fig:scalebar2}
    \end{subfigure}
    \caption{Selected eigenvectors of OA diffusion map on boundaries of LSOA}
    \label{fig:oadiff}
\end{figure}
With the same method as the LSOA diffusion map, we colored the magnitude of eigenvector components on the real map of Bristol city shown in Fig. \ref{fig:oadiff}. Compared to the LSOA diffusion map, the value of entries shows a slightly different scale. However, the patterns of the areas also show a similar pattern to the LSOA diffusion map. Eigenvector 1 still strongly represents the student population, and Eigenvector 2 represents socio-economic deprivation. However, differences in eigenvector entries in a particular area exist between the LSOA diffusion map and the OA diffusion map (LSOA level). We conduct an experiment in which a more effective model for representing deprivation levels can be identified by evaluating the proposed models against actual UK deprivation data.
\section{Quantitative Analysis}
\label{quant}
\begin{figure*}[t]
\centering
    \begin{subfigure}[b]{0.31\linewidth}        
        \includegraphics[width=\textwidth]{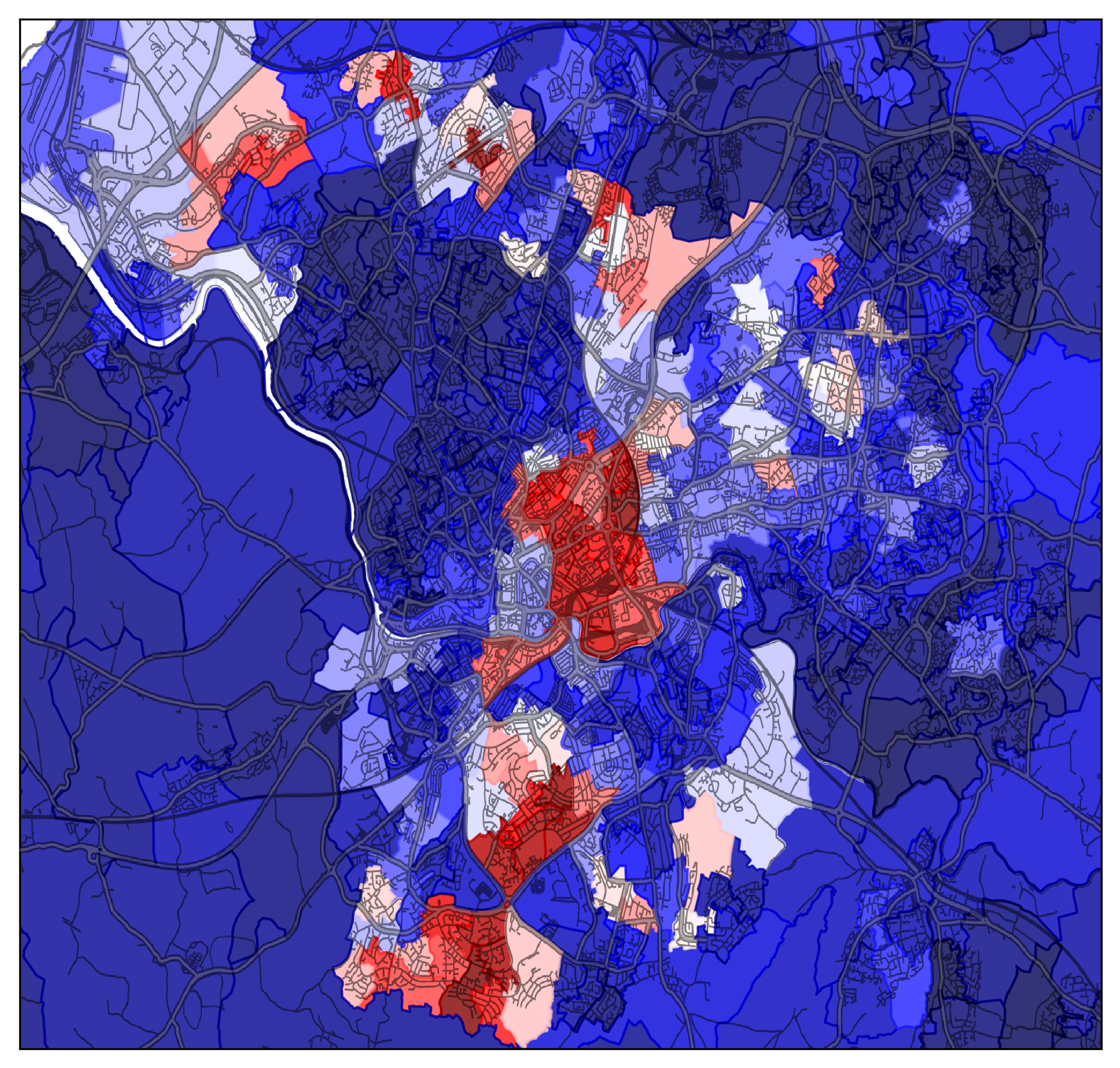}
        \caption[short]{IMD of 2010}
        \label{fig:2010IMD}
    \end{subfigure}
    \begin{subfigure}[b]{0.31\linewidth}        
        \includegraphics[width=\textwidth]{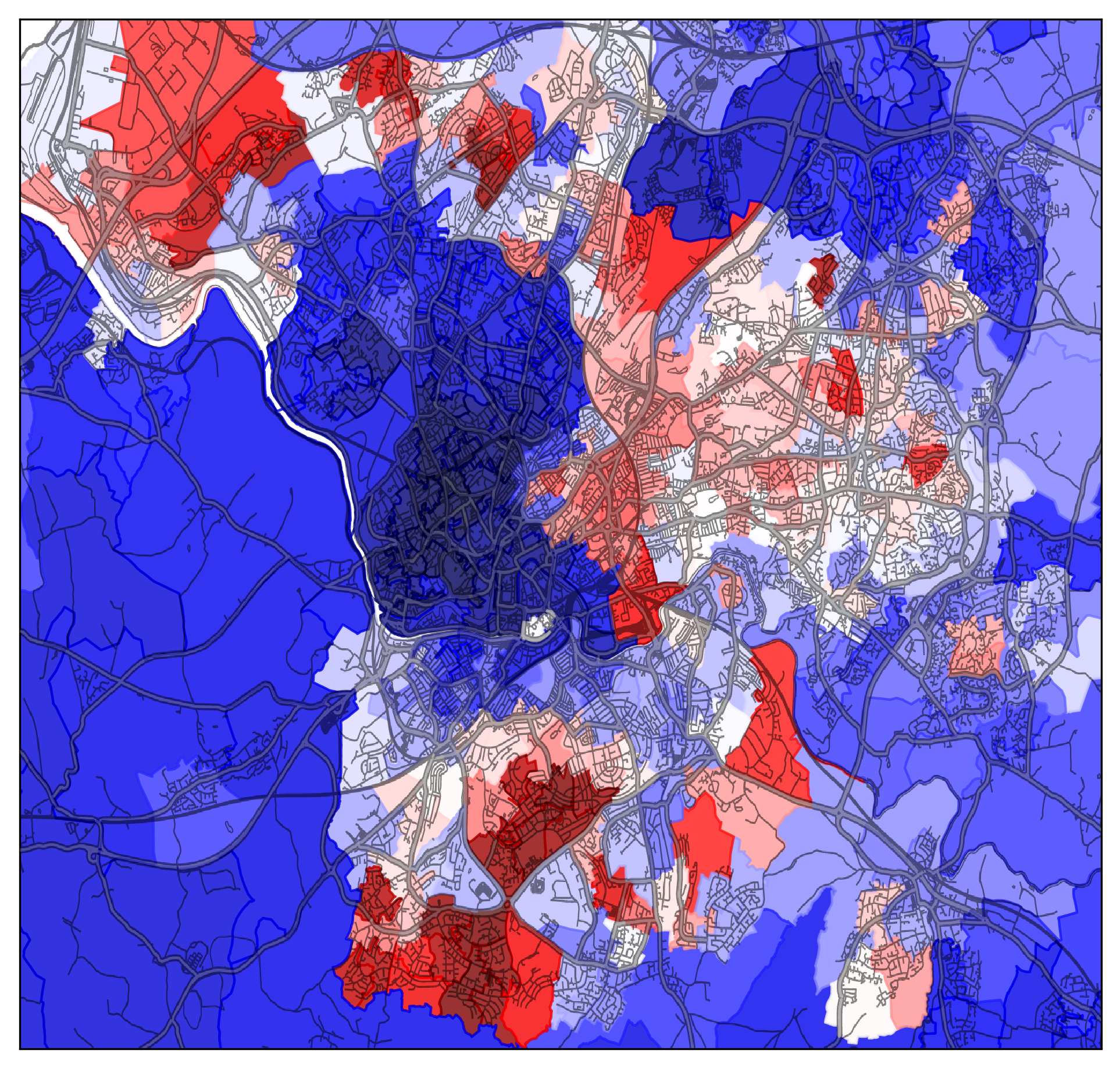}
        \caption[short]{LSOA diffusion map}
        \label{fig:lsoadiffusionmap}
    \end{subfigure}
    \begin{subfigure}[b]{0.31\linewidth}        
        \includegraphics[width=\textwidth]{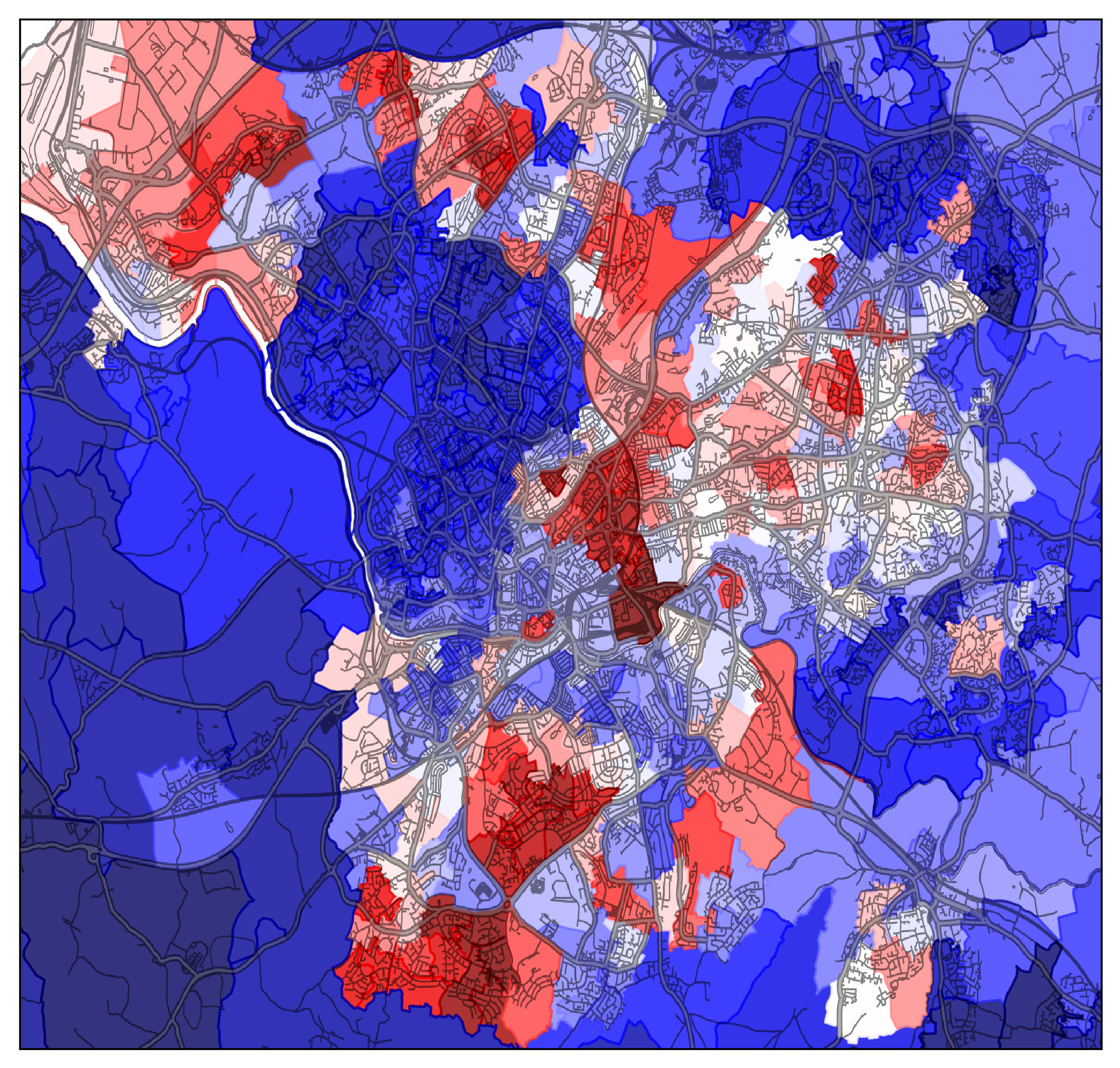}
        \caption[short]{OA diffusion map (LSOA level)}
        \label{fig:oadiffusionmap}
    \end{subfigure}
    \caption{(a) is the map of 2010 IMD; we plot the IMD score of each LSOA. With this map, it is possible to see the patterns of 2010 IMD and compare intuitively with the diffusion map.  For example,  we can see the patterns of red coloured deprived areas are highly similar to the Eigenvector 2 in (b) and (c).}
    \label{fig:oadiff2}
\end{figure*}
To assess whether the diffusion map of census data can represent socio-economic deprivation, and which diffusion map (LSOA diffusion map or OA diffusion map) is a better model, we investigate the correlation between the diffusion maps, along with the 2010 and 2015 UK deprivation data. We compute correlation with the method called the Pearson correlation coefficient, which measures the strength of the association between two variables \cite{benesty2009pearson}.\\\\
Pearson correlation coefficient, commonly referred to as $\rho$, is one of the measures of correlation investigating the linear correlation of two variables. The population Pearson correlation coefficient is the covariance of those two variables divided by the product of their standard deviation. The coefficient range is between $-1$ and 1, where 1 means total positive linear correlation, $-1$ means total negative linear correlation, and 0 indicates no linear correlation\cite{benesty2009pearson}. The correlation between two random variables is defined to be

\begin{equation}
    \rho_{X,Y}=\frac{COV(X,Y)}{\sigma_{X}\sigma_{Y}}
\end{equation}
where
\begin{itemize}
    \item \textit{X, Y}: are the random variables,
    \item \textit{COV}: is the covariance,
    \item $\sigma_{X}$,$\sigma_{Y}$: are the standard deviations of \textit{X} and \textit{Y}, respectively.
\end{itemize}

We use the sample Pearson correlation coefficient, commonly referred to as $\textit{r}$. We substitute the covariances and variances of the sample into the Pearson correlation coefficient formula for a population, to get\\
\begin{equation}
    \textit{r}_{xy}=\frac{\sum^{n}_{i=1}(x_{i}-\overline{x})(y_{i}-\overline{y})}{\sqrt{\sum^{n}_{i=1}(x_{i}-\overline{x})^2}\times\sqrt{\sum^{n}_{i=1}(y_{i}-\overline{y})^2}}
\end{equation}
where:
\begin{itemize}
    \item \textit{$x_{i}$, $y_{i}$}: are the data,
    \item \textit{n}: is the sample size,
    \item $\overline{x}$, $\overline{y}$: are the sample mean of $x_{i}$ and $y_{i}$, respectively.
\end{itemize}

\subsection{2010 deprivation data}
\label{2010_deprived_data}
We choose IMD as an indicator of socio-economic deprivation, representing an overall deprivation score in the UK. With the IMD score, we can analyse numerically the reliability of the model we built by comparing the correlations between vectors of the scores and Eigenvector 2 of each diffusion map. Now, we will investigate IMD and its domains and explore patterns of diffusion maps in a numerical way by computing the Pearson correlation coefficient.


\begin{figure}[h]
    \centering
    \includegraphics[width=\linewidth]{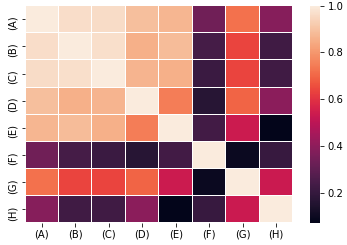}
    \caption{This is a heat map of the Pearson correlation coefficient between IMD and its 7 domains. Each (A) to (H) represents IMD, Income, Employment, Health, Education, Barriers to Housing and services, crime and Living environment domains.}
    \label{fig:heatmap1}
\end{figure}

First, we compute the Pearson correlation coefficient of IMD and its seven domains. Then, we visualise the coefficients by using the heat map to see the correlations intuitively which domains correlate or less correlate with each domain. Shown in Fig. \ref{fig:heatmap1}, for example, four domains (Income, Employment, Health and Education) which profoundly influence the IMD score have a high coefficient, while the rest of the domains (Barriers to housing, Crime, Living environment) relatively have a low coefficient. However, according to Jacob Cohen\cite{cohen1988statistical}, several authors have suggested guidelines to interpret a correlation coefficient, but the criteria are arbitrary (i.e. correlation coefficient 0.8 can be a low correlation for a physical law with high-quality instruments, but it can be a high correlation for social sciences which are largely contributed from complicating factors.). 

Given that our analysis focuses on the relationship with socio-economic deprivation, we have employed a threshold of 0.8 to define and analyse strong correlations. This threshold is chosen to ensure that the identified correlations are both statistically significant and practically meaningful within the context of socio-economic studies, where capturing substantial relationships is crucial for accurate modeling and interpretation.


\begin{figure}[h]
    \centering
    \begin{subfigure}[b]{0.45\linewidth}        
        \centering
        \includegraphics[width=\linewidth]{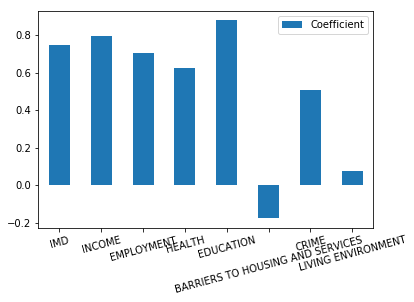}
        \caption{LSOA diffusion map and IMD domains (2010)}
        \label{fig:LSOA2010bar}
    \end{subfigure}
    \begin{subfigure}[b]{0.45\linewidth}        
        \centering
        \includegraphics[width=\linewidth]{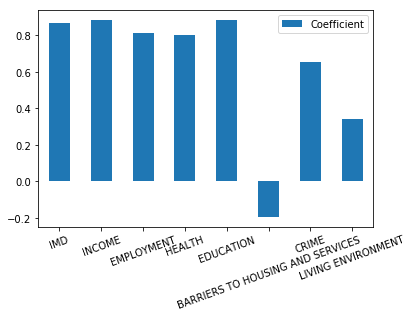}
        \caption{OA diffusion map and IMD domains (2010)}
        \label{fig:OAdiff2010bar}
    \end{subfigure}


    \begin{subfigure}[b]{0.45\linewidth}        
        \centering
        \includegraphics[width=\linewidth]{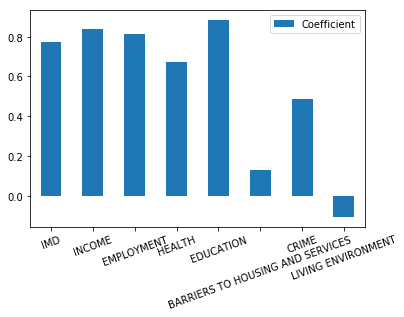}
        \caption{LSOA diffusion map and IMD domains (2015)}
        \label{fig:2015IMDLSOAbar}
    \end{subfigure}
    \begin{subfigure}[b]{0.45\linewidth}        
        \centering
        \includegraphics[width=\linewidth]{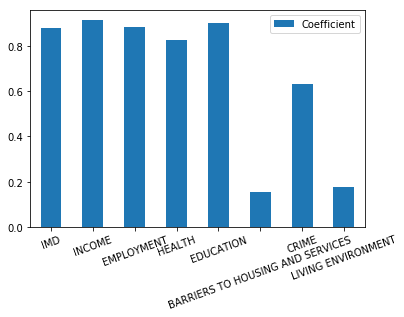}
        \caption{OA diffusion map and IMD domains (2015)}
        \label{fig:2015IMDOAdiffbar}
    \end{subfigure}

    \caption{Pearson's correlation coefficient of IMD domains and Eigenvector 2 diffusion maps for 2010 and 2015}
    \label{fig:combined_pearson}
\end{figure}

We extract the coefficients of IMD and its domains with Eigenvector 2 of diffusion maps and plot them as bar charts in Fig. \ref{fig:combined_pearson}. We intuitively discover that the actual deprived areas (UK deprivation data) and predicted deprived areas (Diffusion map) show similar patterns. These patterns can be explained numerically with the correlation coefficient. From the bar carts, IMD and four domains (Income, Employment, Crime and Living environment) correlate more strongly with the diffusion map than the rest of the domains. The indicators of the census data can explain this result. Census data does not include any information that can indicate barriers to housing, crime, or living environment. (i.e. There are no categories for house prices, criminal records and local environment in the census data.) We expect the domains with low correlation will negatively influence the accuracy of predicting future deprivation, but, at the same time, the Eigenvector 2 of the diffusion map will be a good model for the four highly correlated domains.

In Fig. \ref{fig:combined_pearson}, bar charts \ref{fig:LSOA2010bar} and \ref{fig:OAdiff2010bar} represent the correlation coefficients with IMD score of two diffusion maps: LSOA diffusion map and OA diffusion map in LSOA boundaries. The OA diffusion map (LSOA level) has a stronger correlation with the deprivation data compared to the LSOA diffusion map. Since the OA diffusion map (OA level) considers socio-economic deprivation in the OA level, it is possible to reflect the deprivation level of each area in more detail than the LSOA diffusion map even though the OA diffusion map (LSOA level) has been aggregated to the LSOA level to compare with the IMD data. Therefore, for the 2010 deprivation data, the OA diffusion map (LSOA level) is a better model than the LSOA diffusion map.

\subsection{2015 deprivation data}

Using the same heatmap method, we extract the correlation coefficients between the 2015 IMD, its domains, and Eigenvector 2 of the diffusion maps. Consistent with the results from the 2010 IMD analysis, the Pearson correlation coefficients between the 2015 IMD domains and the OA diffusion map are higher than those for the LSOA diffusion map. Specifically, the average of the top four Pearson correlation coefficients for IMD domains (Income, Employment, Health, and Education) is 0.8021 for the LSOA diffusion map and 0.8788 for the OA diffusion map. These findings suggest that the OA diffusion map based on 2011 census data is more effective at predicting socio-economic deprivation than the LSOA diffusion map.

When comparing the Pearson correlation coefficients between 2010 and 2015, the coefficients for 2015 are generally higher. This difference can be attributed to changes in the census data versions, particularly in the geographical standards such as LSOA boundaries. While the shape of LSOA boundaries remains constant, they may be merged or split depending on significant population changes. As a result, even though the area of an LSOA may expand or contract, it is still possible to compare deprivation levels across LSOAs.


\section{Discussion}
\label{discussion}
\begin{figure*}[t]
    \centering
    \begin{subfigure}[b]{0.45\linewidth}        
        \centering
        \includegraphics[width=\linewidth]{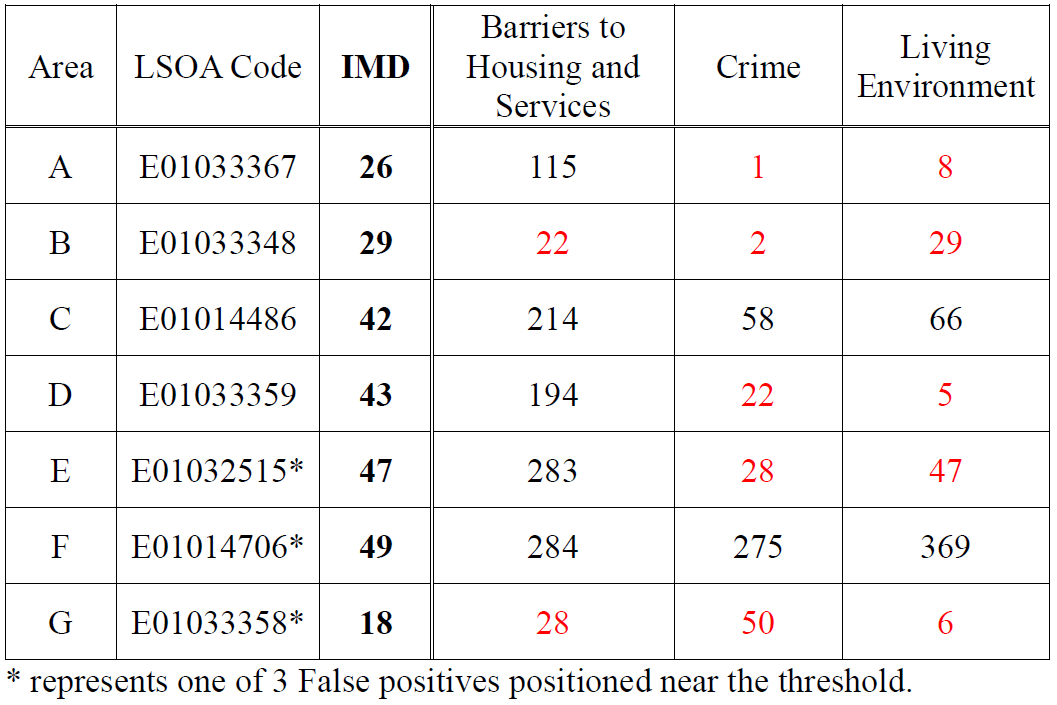}
        \caption{The rank of IMD and 3 weakest correlated IMD domains.}
        \label{fig:7_table}
    \end{subfigure}
    \begin{subfigure}[b]{0.3\linewidth}        
        \centering
        \includegraphics[width=\linewidth]{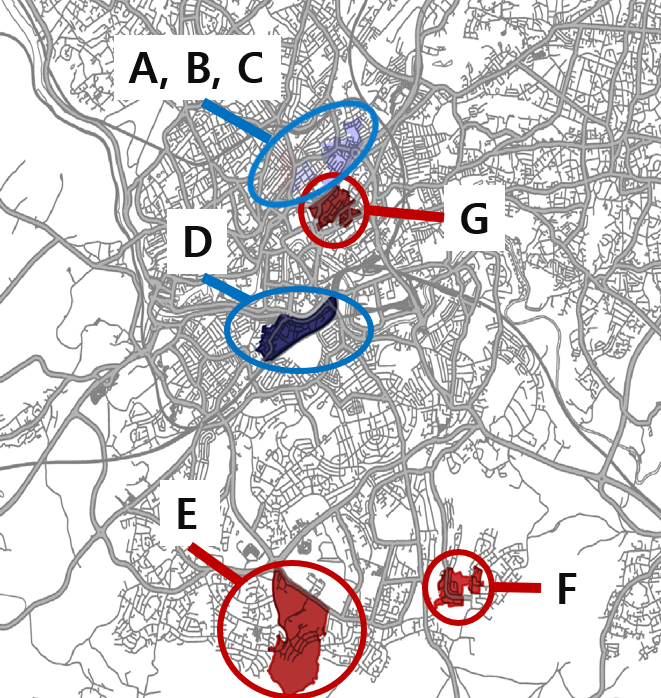}
        \caption{The map of False negatives areas}
        \label{fig:0.0210}
    \end{subfigure}
    \caption{The red coloured values in the table show the domains which are highly ranked in the certain LSOA. Those high ranks are not reflected in the model since those domains have a weak correlation with the model. Red circled regions are the areas where entry of Eigenvector 2 is high enough so positioned near the threshold, while blue circled areas are the areas where entry Eigenvector 2 is too low to cover with decreasing the threshold.}
    \label{fig:table and the map}
\end{figure*}

Eigenvector 2 of the OA diffusion map on LSOA boundaries serves as a good predictor for identifying deprived areas, specifically focusing on assisting government and local authorities in locating these areas with more efficiency. According to the 2015 report from the Department for Communities and Local Government, 3,284 LSOAs in the UK ranked in the top 10 per cent of deprived areas, including 52 in Bristol. Although 7.8 per cent of Bristol's areas face deprivation, it's still below the national threshold, indicating fewer deprived areas compared to other UK cities.
The diffusion map predicted 38 true positive deprived LSOAs and missed 14 false negatives out of the 52 most deprived areas in Bristol. The investigation into the model's inaccuracies focused on two primary hypotheses: first, the structure of computing IMD might affect the outcomes, and second, non-deprived OAs within the falsely predicted LSOAs might be skewing the results. Addressing these hypotheses could improve prediction accuracy.

\subsection{The structure of weighted IMD domains}
\label{sec:weighted_IMD_domains}
First, we consider the false negatives inside the Bristol city: 7 out of 14 false negatives appear inside the city. We assume that the IMD domains which have a weak correlation with the model can influence the entry of Eigenvector 2. For example, if the rank of less correlated IMD domains is high, the ranks of them stand a chance not to be reflected in the model. At the same time, the rank of strongly correlated IMD domains can influence the model too much.

From the result in Fig. \ref{fig:table and the map}, high ranks in weakly correlated IMD domains and low ranks in strongly correlated IMD domains do influence the entry of Eigenventor 2. Therefore, this assumption is the correct reason for the limitation of the model. In contrast, area (C) has low ranks in these domains, so area (C) is not detected for a different reason. The reason is that the Education domain, which is strongly correlated to the model, ranked 152 positioned far from the threshold. This low rank highly reflects the model to predict the deprivation level of the area. (Area (F) is not considered because the predicted deprived level is positioned near the threshold) This project aims to predict the deprivation from the diffusion map; however, we can find the reason for restriction by the location of deprived areas. The following two hypotheses find the reasons inside the diffusion map so we can see the reason from the model and compensate and develop it.

\subsection{Expected deprivation on OA level}
The OA diffusion map (aggregated to LSOA level) is the best model among the diffusion maps to reflect the deprivation in Bristol. This diffusion map is based on the average of the OAs entries located in the certain LSOA. We now investigate whether the Eigenvector 2 entries of OAs in the certain LSOA might be both True positive values and False negative values with the threshold (= 0.02365) of OA diffusion map (LSOA level). 

\begin{figure}[h]
    \centering
    \begin{subfigure}[b]{0.45\linewidth}        
        \centering
        \includegraphics[width=\linewidth]{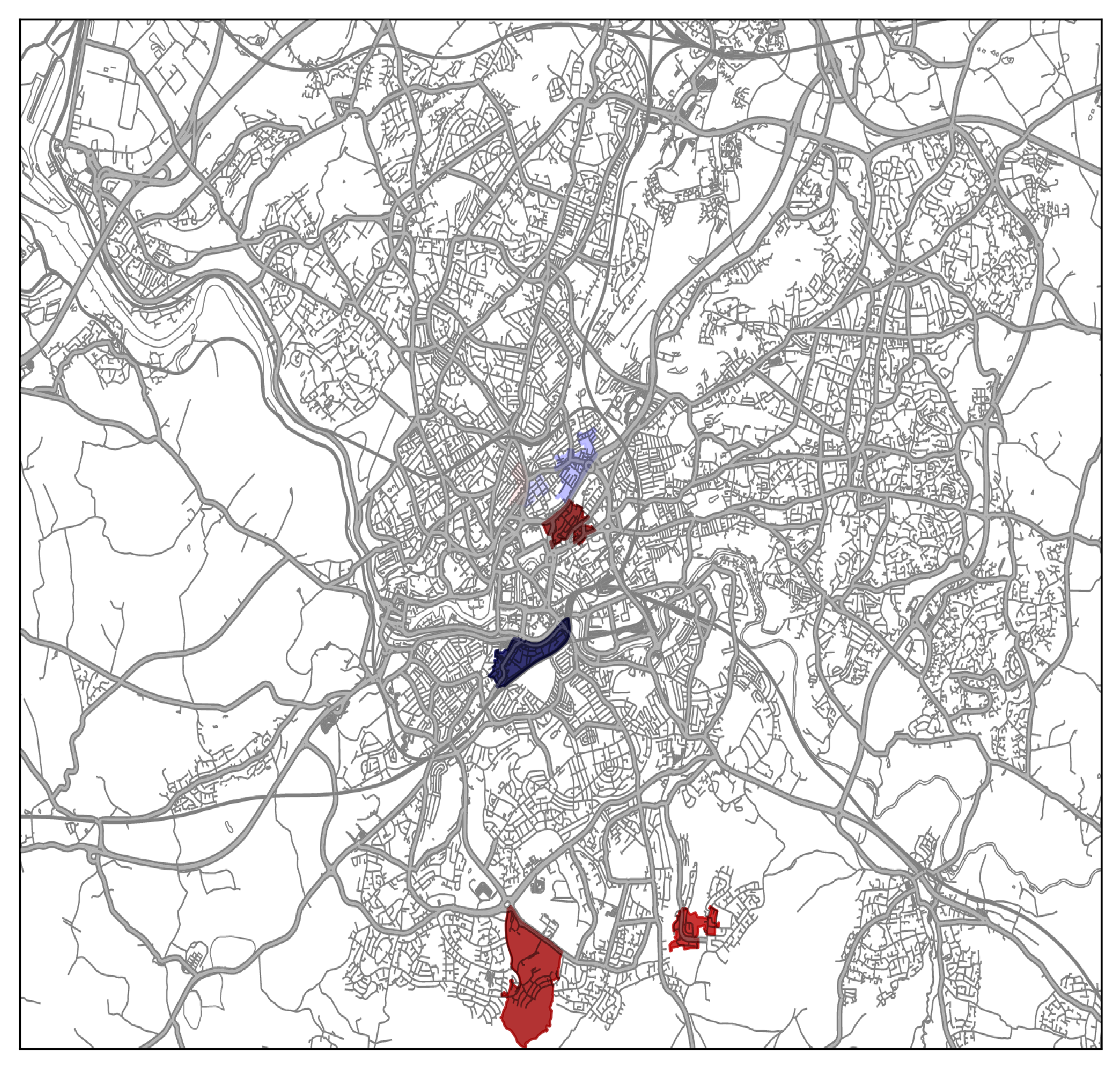}
        \caption{False negatives (LSOA level)}
        \label{fig:FN_LSOA}
    \end{subfigure}
    \begin{subfigure}[b]{0.45\linewidth}        
        \centering
        \includegraphics[width=\linewidth]{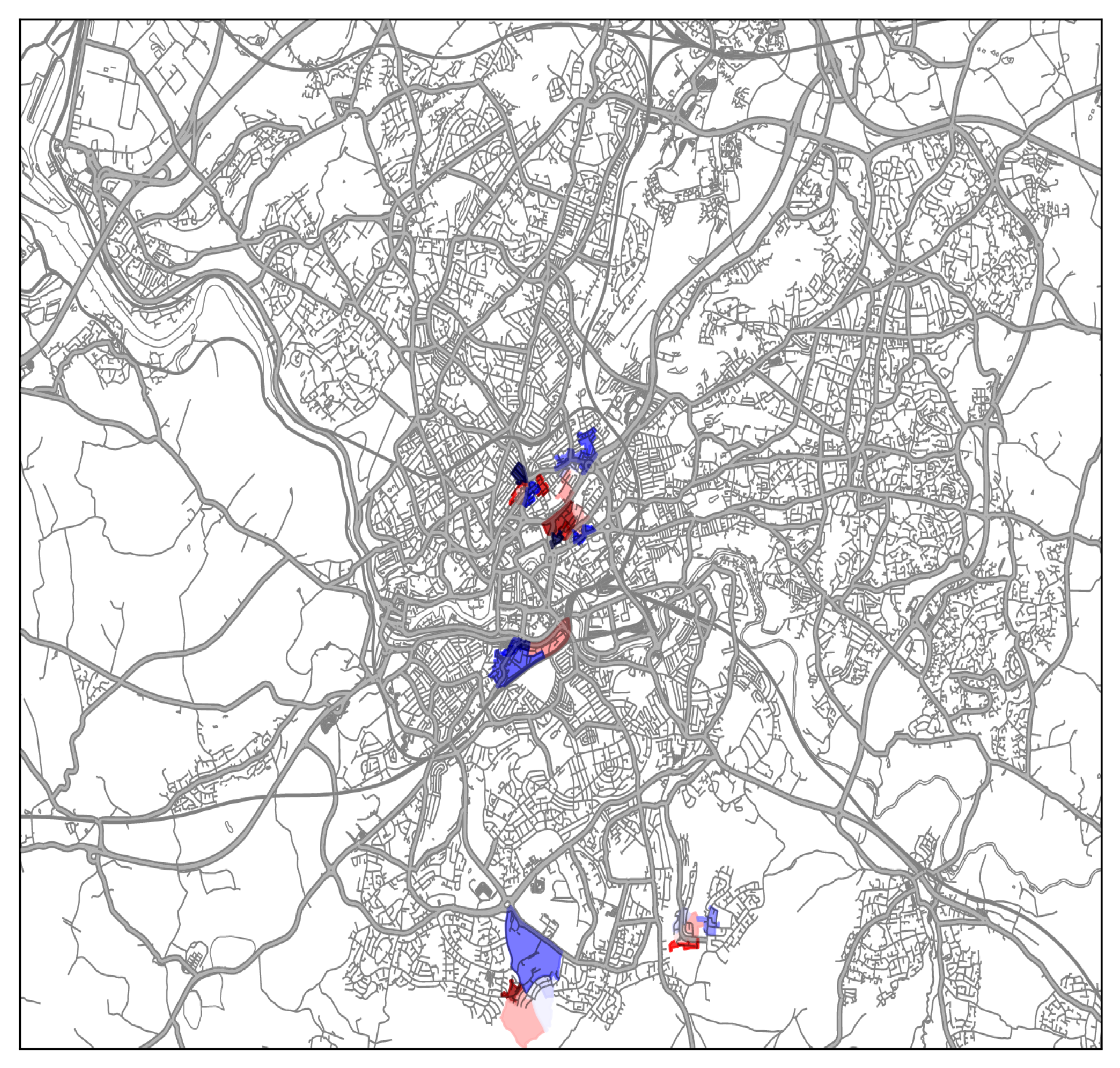}
        \caption{Separated False negatives in OA level}
        \label{fig:FN_OA}
    \end{subfigure}
    \caption{Figure (a) shows the patterns of False negatives on LSOA level. Red areas have a value near the threshold, and blue areas have a low value. These LSOAs are separated into OA level and allocated their values from OA diffusion map (OA level). In (b), each LSOA includes True positive and False negative values of OAs.}
    \label{fig:FNFNFNFNFNFNFN}
\end{figure}

84 OAs belongs to these 14 False negative LSOAs. We find that 20 out of 84 OAs have True positive values. On the other hand, 64 OAs are False negatives and these OAs influence the entry of the 14 LSOAs to be False negatives. To numerically investigate the False negatives, we plot their entry of OAs and LSOAs.

\begin{figure}[h]
\centering
\includegraphics[width=\linewidth]{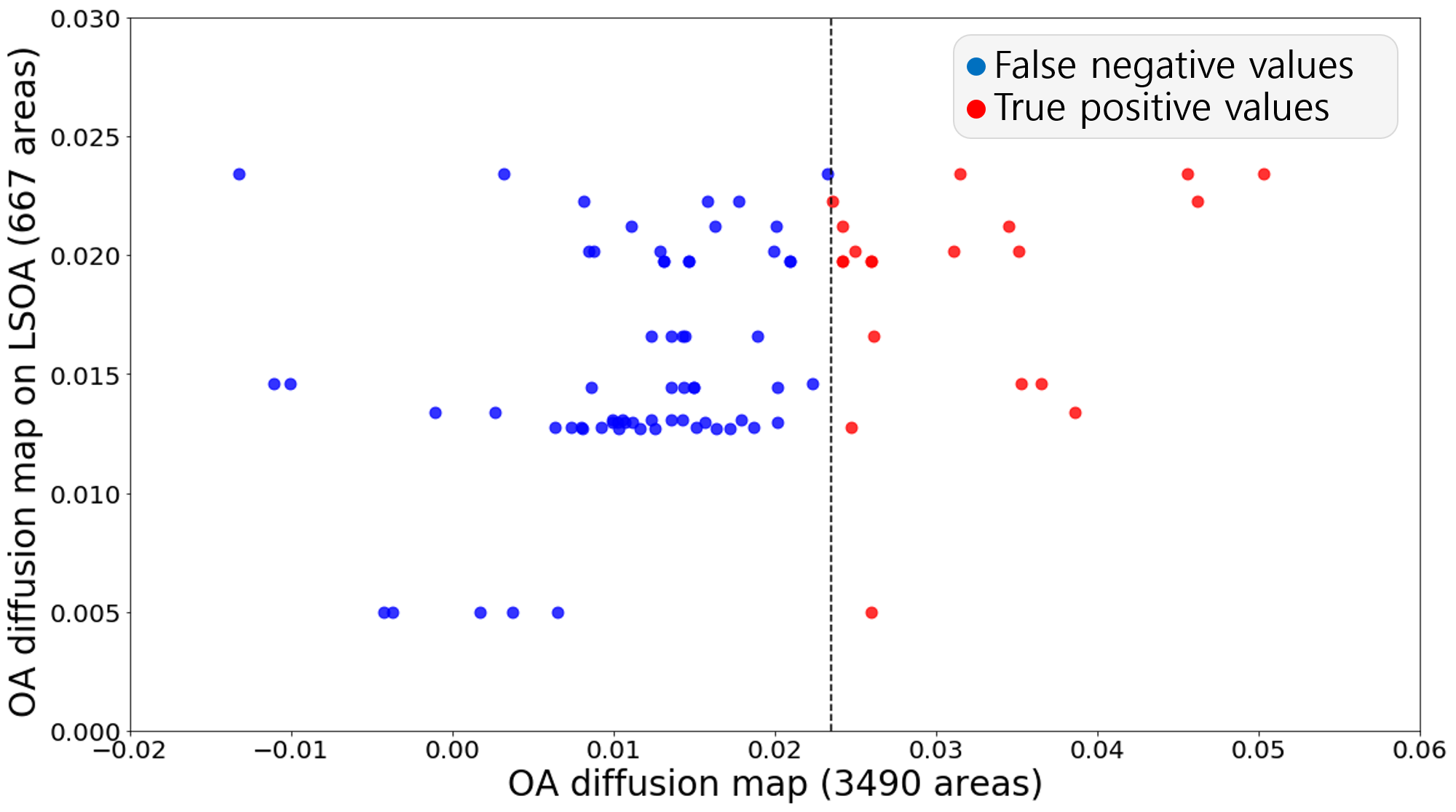}
\caption{We separate the 14 False negatives LSOA into OA level and plot their Eigenvector 2 component values. In the Figure, there exist approximately 24 per cent of OAs that are over the threshold. }
\label{fig:graph_OA_LSOA}
\end{figure}

As shown in Fig. \ref{fig:graph_OA_LSOA}, even though plenty of OAs which have entries with high values of Eigenvector 2 exist, LSOAs which include these OAs are classified as False negatives. Therefore, if we focus on deprivation level in OA rather than LSOA, it is possible to put the deprivation level in smaller areas into practice. 
\section{Conclusion}
\label{conc}
In this research, we crafted a model for socio-economic deprivation using a diffusion map informed by high-dimensional census information. We evaluated two models: one at the Output Area (OA) level and another at the Lower Layer Super Output Area (LSOA) level. Results indicated that the OA diffusion map applied to LSOA boundaries showed a stronger correlation with the UK Index of Multiple Deprivation (IMD) data from 2010 and 2015. Although the model effectively captured domains such as income, employment and education, it exhibited a weaker correlation with housing barriers, crime, and the living environment. By concentrating on Bristol's most deprived areas, we pinpointed 52 regions, with the OA diffusion map accurately forecasting 38. The remaining 14 regions (false negatives) likely resulted from weaker domain correlations and the averaging at the OA level. Despite these constraints, our model exhibited substantial predictive capability overall. Future research could target the False negatives and refine accuracy for specific IMD domains. This method, however, provides a valuable mechanism for detecting deprivation at a more granular resolution (OA level), potentially enhancing governmental resource distribution efficiency.
\section*{Acknowledgment}
The author thanks Prof. Thilo Gross for supervising the project at the University of Bristol and Dr. Ed Barter for engaging in insightful discussions about diffusion maps and providing valuable guidance throughout the project.
\bibliographystyle{IEEEtran}
\bibliography{reference.bib}

\end{document}